\documentclass[10pt,a4paper]{article}
\usepackage{authblk}
\usepackage{multibib}
\usepackage{lrec}
\usepackage{times}
\usepackage{url}
\usepackage{latexsym}
\usepackage{graphicx}

\usepackage{multirow}
\usepackage{graphicx}

\title{Events Beyond \textsc{ACE}: Curated Training for Events}

\author[1]{Ryan Gabbard}
\author[2]{Jay DeYoung}
\author[1]{Marjorie Freedman}
\affil[1]{Information Sciences Institute, University of Southern
    California
\\ \texttt{gabbard@isi.edu,mrf@isi.edu} }
\affil[2]{Raytheon BBN Technologies
\texttt{\{jay.deyoung@raytheon.com\}}}

\name{Ryan Gabbard, Jay DeYoung, Marjorie Freedman\thanks{Work was
performed when
Gabbard and Freedman were at BBN}}
\address{USC Information Sciences Insitute, Raytheon BBN Technologies  \\
\texttt{gabbard@isi.edu}, \texttt{jay.deyoung@raytheon.com},
\texttt{mrf@isi.edu}}

\date{}

\abstract{
We explore a human-driven approach to annotation, \emph{curated training}
(\textsc{ct}), in which annotation is framed as teaching the system by using
interactive search to identify informative snippets of text to annotate,
unlike traditional approaches which either annotate preselected text or use
active learning. A trained annotator performed 80 hours of \textsc{ct} for
the thirty event types of the \textsc{nist} \textsc{tac kbp} Event Argument
Extraction evaluation. Combining this annotation with \textsc{ace} results
in a 6\% reduction in error and the learning curve of \textsc{ct} plateaus
more slowly than for full-document annotation.  3 NLP researchers performed
\textsc{ct} for one event type and showed much sharper learning curves with
all three exceeding \textsc{ace} performance in less than ninety minutes,
suggesting that \textsc{ct} can provide further benefits when the annotator
deeply understands the system.
}

\begin{document}
\maketitleabstract
\section{Introduction}

Identifying events
from a given ontology in text and locating their arguments is an especially challenging task because events vary widely in their textual realizations and their arguments
are often spread across multiple clauses or sentences.  Most event research
has been in the context of the 2005 NIST
Automatic Content Extraction (\textsc{ace}) sentence-level event mention task
\cite{ace2005}, which also provides the standard corpus. Recently, \textsc{tac kbp} has introduced document-level event argument extraction
shared tasks for 2014 and 2015 (\textsc{kbp ea}).

Progress on events since ACE has been limited.   Most subsequent work has tried
improve performance through the use of more complex inference \cite{li13}, by
transductively drawing
on outside sources of information, or both \cite{ji08,hong11}.
Such approaches have produced modest reductions in error over a pipeline of simple classifiers trained on ACE.

In our efforts to improve on the \textsc{kbp ea} 2014 systems, we  were
stymied by a lack of data, especially for rarer
event types.  Ten of the 33 event types have fewer than 25 training examples in \textsc{ace},
and even for more frequent events, many trigger words and classes of arguments occurred only once.  
Furthermore, the 2015 task would include new argument types.  These problems motivated the following question: 
 (a) are we at a plateau in the performance vs. annotation time curve?
(b) is there an viable alternative to full-document annotation, especially for rarer event types? (c) 
for novel event types or languages, how quickly can a useful event model be trained? 

In traditional annotation, a static corpus selected 
to be rich in the target event types is annotated.
Active learning augments existing training 
data by having a human oracle annotate system queries (or features \cite{settles11}).   
We explored a novel form of
annotation, \emph{curated training} (\textsc{ct}), in which \emph{teachers} (annotators) actively
seek out informative training examples.

\section{Curated Training}

In \textsc{ct} the teacher created a prioritized \textsc{indicator list} of 
words and phrases
 which could indicate a target event's presence.  Given a tool with a search box, a document
list, and a document text pane,  teachers searched\footnote{over
Gigaword 5 \cite{gigaword5} using Indri \cite{indri}} for
indicators in priority order and annotated ten documents each.  
On loading a document, they
used their browser's search to locate a single sentence containing the indicator.

If the  sentence mentioned multiple instances of the target event or was
unclear, it was skipped.  If it contained no
mention of the event, they marked it \textsc{negative}.\footnote{Negated,
future, and hypothetical events were all considered mentions of an event, not \textsc{negatives}}
 Otherwise, they (a) marked the sentence as \textsc{event-present};  (b)
 applied the \textsc{anchor} annotation to the tokens\footnote{if there were no anchors, the document was
 skipped} whose presence makes the presence of the event
 likely; (c) marked each argument span within the selected
 sentence; and (d) marked any other spans they thought might be `educational'
 as \textsc{interesting}.  (d) was also done for \textsc{negative}
 sentences.

Teachers were permitted to annotate extra documents if
an indicator seemed ambiguous.  They looked very briefly (2-3 seconds) in the context of selected sentences to
see if there were additional informative instances to annotate.
If any non-indicator anchor was marked, it was added to the indicator list with high priority.
The process was repeated for four hours or until the teacher
felt additional \textsc{ct} would not be useful.\footnote{See curves in
Figure \ref{fig:per-event}. In many cases annotation appears to terminate
early because annotators had no way of tracking when they hit exactly four
hours.}  

\subsection{Data Gathered}

 We recruited three teachers without NLP backgrounds but with annotation experience.
We consider here only the teacher (\textsc{a}) who completed all event types in
time for assessment.  Teacher \textsc{a}
averaged seven minutes brainstorming indicators and produced 6,205 event
presence, 5137 negative, and 13,459 argument annotations.  Every teacher action was time-stamped. For analysis,  we updated the timestamps to
 remove breaks longer than two minutes.\footnote{Annotators were never to
spend more than 2-3 seconds on any decision } 

Since the \textsc{ct} was stored as character offsets, we aligned it to parses  
to get \textsc{ace}-style event mentions. For training our argument
attachment models, we omit any event mentions where any annotation
failed to project.  Projecting Teacher \textsc{a}'s
data produced 5792 event mentions for trigger training, 5221 
for argument training, and 4,954 negatives.\footnote{c.f. roughly 5,300 event mentions in \textsc{ace}} 

\section{Evaluation}

Our target evaluation task is \textsc{kbp-ea} \cite{ea-2014} which requires 
mapping a document to a set of $(t,r,e,m)$  tuples
indicating that an entity $e$ plays the role $r$ in an event of type $t$
with realis $m$. Scoring is
 F1 over these tuples.\footnote{The details of the scoring are given
in \cite{ea-2014}.} We evalute over the
the 2014 newswire evaluation corpus \cite{ldc-2014-ea-eval-corpus} using 
the scorer\footnote{https://github.com/isi-nlp/tac-kbp-eal}
on the evaluation key augmented with assessments by Teacher \textsc{a} of responses from our system not found therein.\footnote{
The evaluation answer key had assessments of all 2014 system response and those of an an LDC annotator operating under significant time-pressure
(thirty minutes per document)} To 
focus on event detection and argument attachment, we enabled the
\texttt{neutralizeRealis} and \texttt{attemptToNeutralizeCoref} scorer
options.
\subsection{Baseline}
The highest-performing system in \textsc{kbp ea} 2014, \textsc{bbn1}, ran a pipeline
of four log-linear classifiers (trigger detection, argument attachment, genericity assignment, and a trigger-less argument model)
in a high-recall mode which output all event mentions and arguments
scoring above 10\% probability.  This output was fed into a series of 
inference rules and a score was computed based on the sub-model scores 
and the inference rules applied \cite{bbn-2014-ea-system}.

We used this evaluation system for the experiments in this paper 
with two changes.  First,
\textsc{bbn1} used a multi-class model for trigger detection, while we
use one binary model per event type because with \textsc{ct}
each type has a different set of negative examples.  Second, we omitted the
`trigger-less'  argument classifier for simplicity.
This version, \textsc{baseline}, lags \textsc{bbn1}'s
performance by 0.8 F1 but outperforms all other 2014 evaluation systems by
a large margin.

To compare against full document
annotation, we needed to estimate how long the event-only portion of
\textsc{ace}
annotation took.\footnote{Excluding coreference, etc.}  The LDC\footnote{personal communication} ventured a rough estimate of 1500 words per hour (about twenty
minutes per \textsc{ace}
document).  The LDC human annotator in \textsc{kbp-ea} 2014 was allocated
thirty minutes per document \cite{kbp-ea-overview}.   We use the former
estimate.  To estimate performance
with a fraction of \textsc{ace}, we used the first $n$\% documents as needed.

\section{Analysis}
\begin{figure}[tbp]
\includegraphics[width=\linewidth]{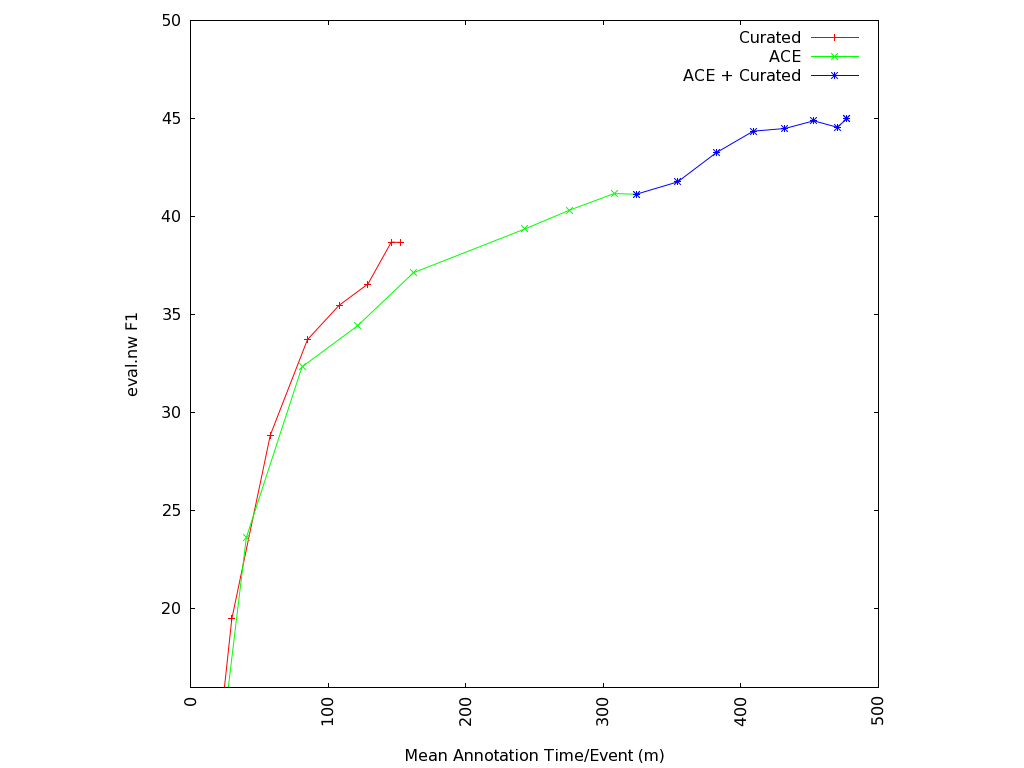}
\caption{Performance vs. mean time per event}
\label{fig:performance-vs-time}
\end{figure}

\begin{figure*}[tbp]
\includegraphics[width=\textwidth,height=\textheight]{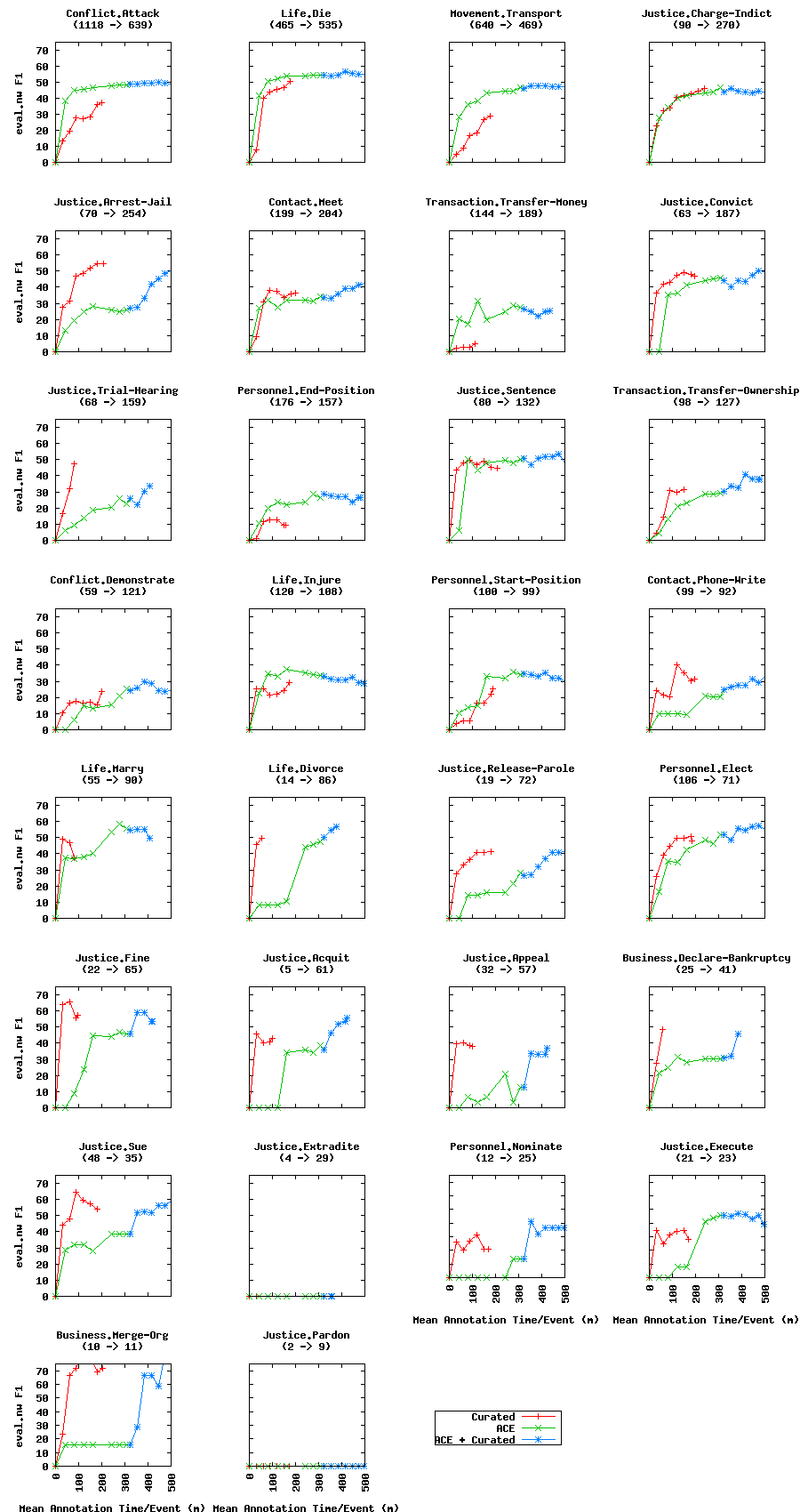}
\caption{Performance vs. average annotation time per event on a per-event
basis.  $x \rightarrow y$ indicates there were $x$ event mentions of training for
this type in \textsc{ace} and $y$ argument tuples for it in the evaluation
set.}
\label{fig:per-event}
\end{figure*}

In aggregate \textsc{ct}'s performance 
closely tracked \textsc{ace} for small amounts of mean annotation time per event (Figure
\ref{fig:performance-vs-time}). However,  the performance
of \textsc{ct} plateaus more slowly than ACE,  beginning to
diverge around ninety minutes per event, and continuing
to increase sharply at the end of our annotation, leaving unclear what
the potential performance of the technique is.  When added to \textsc{ace}, the \textsc{ct} improves performance
somewhat, reducing error of P/R/F  1\%/5\%/6\% at ninety
minutes per event before plateauing.  
\textsc{ct} has a substantial advantage over \textsc{ace} for event types which
are rare in \textsc{ace}, but lags significantly for event
types abundant in \textsc{ace} (Figure \ref{fig:per-event}).\footnote{The anomalously poor performance on
\textsc{transaction.transfer-money} is due to a bug.} 

\begin{figure}[tbp]
\includegraphics[width=\linewidth]{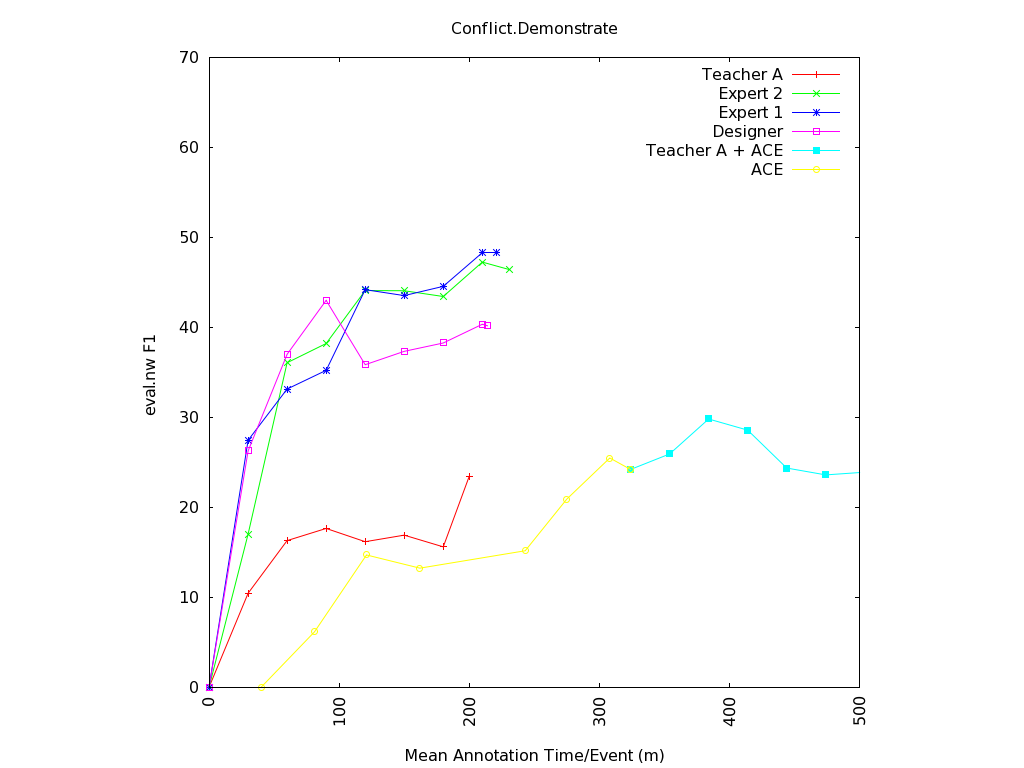}
\caption{Teacher \textsc{a} vs NLP experts on \textsc{conflict.demonstrate}}
\label{fig:experts}
\end{figure}

The annotation tool designer and two other NLP experts also 
did \textsc{ct} for \textsc{conflict.demonstrate} (Figure \ref{fig:experts}; Table \ref{tab:experts}).  
All experts significantly outperformed Teacher \textsc{a} and \textsc{ace} in terms of F1.  In two cases this 
is because the experts sacrificed precision for recall.  The second expert matched Teacher \textsc{a}'s
precision with much higher recall.  Annotators varied widely in the volume of their annotation and indicator searches,
but this did not have a clear relationship to performance.
  
\begin{table}[tbp]

\begin{tabular}{r|ccccc}
& Tch. \textsc{a} 
& \textsc{Designer} & \textsc{Exp. 1}  & \textsc{Exp. 2} 
\\
Words &  13k & 21k & 13k & 20k \\
Doc.s & 256 & 466 & 165 & 334 \\
Searches & 28 & 75 & 23 & 37 \\
Prec. & 71 & 42 &  56 & 71 \\
Rec. & 14 & 38 & 43 & 34 \\
\end{tabular}

\caption{Teacher \textsc{a} vs NLP experts on \textsc{conflict.demonstrate}}
\label {tab:experts}
\end{table}

\subsection{Possible Confounding Factors}
Because Teacher \textsc{a} both provided \textsc{ct} and did the output
assessment, improvements may reflect the system learning
their biases. We controlled for this somewhat by having
Teacher \textsc{B} dual-assess several hundred responses,
resulting in encouraging agreement rates of 95\% for event presence, 98\%
for role selection, and 98\% for argument assessment.\footnote{\textsc{aet}, \textsc{aer}, and
\textsc{bf} in \textsc{kbp ea} terms \cite{kbp-ea-assessment-guidelines}} For some events, the guidelines changed from \textsc{ace} to
\textsc{kbp ea} 2014 by eliminating `trumping' rules and expanding
allowable inference, which could also
account for some improvement.   If either of these were significant factors, it would suggest that \textsc{ct} may be a
useful tool for retargetting systems to new, related tasks.



\section{Acknowledgements}

Thanks to Elizabeth Boschee and Dan Wholey for doing annotation. 
This research was developed with
    funding from the Defense Advanced Research Projects Agency (DARPA). The views, opinions, and/or
    findings expressed are those of the author(s) and should not be
    interpreted as representing the official views or policies of the
Department of Defense or the U.S. Government. Distribution
`A': Approved For Public Release, Distribution Unlimited.

\section{Bibliographical References}
\bibliographystyle{lrec-fix-capitals}
\bibliography{curated-training}

\end{document}